\documentclass[letterpaper]{article}

\usepackage[eatrack]{log_2022}  % for accepted extended abstracts

\usepackage{adjustbox}
\usepackage{amsfonts}           % blackboard math symbols
\usepackage{duckuments}         % sample images
\usepackage{bm}
\usepackage{xcolor}
\usepackage[backref=page]{hyperref}
\usepackage[numbers,compress,sort]{natbib}

\usepackage{amsmath,amsfonts,bm}

\def\eqref#1{equation~\ref{#1}}
\def\1{\bm{1}}

\def\mX{{\bm{X}}}

\DeclareMathAlphabet{\mathsfit}{\encodingdefault}{\sfdefault}{m}{sl}
\SetMathAlphabet{\mathsfit}{bold}{\encodingdefault}{\sfdefault}{bx}{n}

\def\gG{{\mathcal{G}}}

\def\sV{{\mathbb{V}}}

\usepackage{booktabs}           % professional-quality tables
\usepackage{graphicx}           % figures
\usepackage{multirow}           % tabular cells spanning multiple
\usepackage{cleveref}
\usepackage{adjustbox}
\usepackage{enumitem}

\title{Revisiting Embeddings for Graph Neural Networks}

\author[S. Purchase et al.]{%
Skye Purchase\\
\institute{University of Cambridge}\\
\email{atp45@cam.ac.uk} \And
Aaron Zhao\\
\institute{University of Cambridge,Imperial College London}\\
\email{a.zhao@imperial.ac.uk}\And
Robert Mullins\\
\institute{University of Cambridge}\\
\email{robert.mullins@cl.cam.ac.uk}
}

\begin{document}

\maketitle

\begin{abstract}

%Overview of the paper: Current method of GNN, the shortcomings of this approach, the method we used to analyse these shortcomings, our findings and the conclusion/further questions.

Current graph representation learning techniques use Graph Neural Networks (GNNs) to extract features from dataset embeddings.
In this work, we examine the quality of these embeddings and assess how changing them can affect the accuracy of GNNs.
We explore different embedding extraction techniques for both images and texts; and find that \emph{the performance of different GNN architectures is dependent on the embedding style used}. 
We see a prevalence of bag of words (BoW) embeddings and text classification tasks in available graph datasets. Given the impact embeddings has on GNN performance. this leads to a phenomenon that \textit{GNNs being optimised for BoW vectors}.
%In addition, we only see an improvement in accuracy from some GNN models compared to the accuracy of models trained from scratch or fine-tuned on the underlying data without utilising the graph connections. 
% As an alternative, we propose \textbf{\emph{Graph-connected Network (GraNet)}} layers 
% to better leverage existing unconnected models within a GNN. Existing language and vision models are thus improved by
% allowing neighbourhood aggregation. This gives a chance for the model to use pre-trained weights, if possible, 
% and we demonstrate that this approach improves the accuracy compared to traditional GNNs: on Flickr\_v2, GraNet beats GAT2 and GraphSAGE by $7.7\%$ and $1.7\%$ respectively.

\end{abstract}

\section{Introduction}

%What are we trying to do and why is it relevant.

% Current Graph Neural Network approaches use a limited set of node embeddings when evaluating performance

%% Current Graph Neural Networks are evaluated on a limited set of datasets.
%% The datasets based on real world data require preprocessing to a computationally digestable format.
%% We refer to this digestable formats as embeddings.

% Graph Neural Networks (GNNs) have been successful on a wide array of applications ranging from computational biology \cite{} to social networks \cite{}. However, 
Current advancements in Graph Neural Networks (GNNs) are being evaluated on a small range of tasks and accompanying datasets. Though these datasets are sourced from different domains, they require preprocessing the raw data into a computationally digestable format to be usable by GNNs, referred to as \textit{embeddings}. 
%This computational digestable format is what we refer to as \textit{embeddings} and 
In this work we focus on node classification and thus \textit{node embeddings}.

% The most common node embedding is not widely applicable to all types of data

%% The most common datasets used in GNN evaluation focus on simple text classification.
%% To convert raw text into a digestable format the cannonical approach is bag of words.
%% This means that generally graph neural networks are being optimised primarily for bag of word vectors.

Common node classification datasets \cite{shchur2018pitfalls,zeng2019graphsaint,Hamilton:2017tp,Kipf:2017tc} focus on text classification with the primary node embedding being Bag of Words (BoW). Though this is a suitable method for text, this results in current GNNs being optimised to BoW. Equally, this form of node embedding is not always applicable, image data for example, and so GNNs are only being optimised for limited forms of data, mainly text.
Existing literature has focused on the shortcomings of GNN training and the effect that the dataset can have on the model performance \cite{shchur2018pitfalls}, but there is no comment on \textbf{how different node data preprocessing may affect performance}. 
To demonstrate this problem we introduce three new datasets as alterations of existing datasets that are commonly used in literature. 
%%% These alterations retain the underlying graph structure of the existing datasets to avoid our results being dependant on the graph structure.
% These use a variety of embeddings rather than the common embedding technique
%% We build the datasets from source providing raw data as well as embeddings.
%% The embeddings cover a range of techniques focusing on modern neural networks to encode better embeddings.
%% The standard bag of word embedding is still included where relevant.
%% But we do see that BoW is not always sensible.
Each dataset is accompanied by a set of node embeddings.
%%% , these cover a range of techniques focusing on using modern neural networks to provide better node embeddings. 
%Where relevant we include the standard BoW as we do not believe this standard is poor but rather that this approach is not always applicable. 
% We use modern approaches to creating our embeddings
%How do we verify our hypothesis
% We see that the performance of the same GNN depends on the embedding used
%% When looking at the performance of a GNN model across different embeddings we see significant variation.
%% Maintaining the same data source, data splits, GNN architecture and hyperparameters.
%% This suggests the performance of the model depends on the embedding.
To evaluate the effect of node embeddings on GNN performance we train and test standard GNN archictures: Graph Convolution Network \cite{Kipf:2017tc}, Graph Attention Network\cite{Velickovic:2018we} (with GATv2 \cite{Brody2021GATv2}), and GraphSAGE \cite{Hamilton:2017tp} with two different samplers. 
%%% Across the datasets and embeddings the architectures and hyperparameters are kept consistent. We maintain the same database splits, comprising of train, test and validation, across the different embeddings.
% We see that the best performing GNN is not the same for each embedding
%% An important aspect of evaluating GNNs is comparing to prior models.
%% When comparing the ranking of models across two different embeddings we see a shuffle in order.
%%% When using this setup, where we endeavour to only vary the embedding. 
For these models we find that their performance and relative rank is dependant on the embeddings used. 
%%% This suggests that beyond embeddings effecting how well a GNN performs they also effect which GNN architecture is best suited.
% We see that in some cases the status quo is optimal but the status quo is not always optimal
%% We do see that in the case of simple text classification BoW is optimal.
%% We do not look into whether this is the optimal for all cases where it is suitable.
%% But again it is important to note BoW is not always possible.
In this work we make the following contributions:

\begin{itemize}
    \item We put forward three new datasets and a rich set of accompanying embeddings to better test the performance of GNNs. 
    \item We demonstrate that GNN performance depends on the embedding used. The choice of embedding provides large variance and prevents a fair comparison of different architectures.
    \item We demonstrate that current GNN architecture design overfits to limited styles of embedding.
\end{itemize}
\section{Background and Related Work}

%Any notation and formalism

% Introduction to Graph Data

%\textbf{Graph Data} Let $\gG(\sV, \mathcal{E}, \mX)$ denote a graph where $\sV = \{\vv_1, \vv_2, ..., \vv_n\}$ is the set of nodes in the graph, $\mathcal{E}$ is the set of edges between nodes in $\sV$ such that $\ve_{i,j} \in \mathcal{E}$ denotes a directed connection from node $\vv_i$ to node $\vv_j$, and $\mX$ is the raw node data of the graph.
%We say each node $\vv_i$ has a neighbourhood $\sN_i$ such that $\vv_j \in \sN_i \iff \ve_{j,i} \in \mathcal{E}$ and we say that $\vv_j$ is a neighbour node to $\vv_i$. Where $\mX$ is the raw node data matrix where $\mX_{:, i} = \vx_i$ where $\vx_i$ is the feature vector for node $\vv_i$.

% Introduction to Embeddings

%% Given the raw data in graph G, X, an embedding is some transformation of this data, X_e
%% X_e is a more compact feature space better suited for GNN training
%% This transformation, where applicable, can be represented as some embedding function f_e

% \subsection{Node Embedding} 
Given a graph, $\gG(\sV, \mathcal{E}, \mX)$, with raw node data, $\mX$, there exists a transformation function, $f_e$, to project the raw data to a more compact feature space, $\mX_e$, such that $\mX_e = f_e(\mX)$. A GNN then trains and is evaluated on this transformed node data rather than the original raw data. A \textit{(dataset ($\gG$), embedding ($f_e$))} pair is a specific graph, $\gG$, with a specific embedding function, $f_e$.

%% In the rest of the paper we will discuss (dataset, embedding) pairs to mean a specific graph dataset D with graph G and an embedding of the graph data, X_e

%Problem setting and concepts/prior work required to understand this approach

% Description of the models that we compare against

%% This paper uses the standard GCN, GAT and GraphSAGE setup to demonstrate the effect of embeddings on model performance
% Overview of these techniques

This work focuses on the standard set of Graph Neural Networks (GNNs): Graph Convolution Network (GCN) \cite{Kipf:2017tc}, Graph Attention Network (GAT \cite{,Velickovic:2018we} and GATv2 \cite{Brody2021GATv2}) and graphSAGE \cite{Hamilton:2017tp}.
%\citet{Kipf:2017tc} introduce GCN (Graph Convolutional Networks) -- a method of applying convolutional layers from CNNs to graph neural networks. It focuses on spectral filters applied to the whole graph structure rather than at the node level.
%\citet{Hamilton:2017tp} introduce the GraphSAGE model which builds on prior work from GCN focusing on individual node representations. This gives rise to the iterative message passing process on the node level. Though simpler than newer models we find that this approach, when given the right embedding style, can outperform some recently published GNNs.
%\citet{Velickovic:2018we} introduce the idea of graph attention which alters how a node aggregates its neighbours representation. This adds an additional attention mechanism to discern which aspects of the node representations in a nodes neighbourhood are important at a given layer. 
%\citet{Brody2021GATv2} provide a more attentive version based on the graph attention system introduced in \citet{Velickovic:2018we}. We base the graph attention mechanism used in our GraNet models on this improved version of graph attention. We provide both versions of graph attention in our results to compare to our new approach.
% Look at the datasets that these papers use as baselines
%% These three papers focus on a small subset of datasets: Citation networks, Reddit, Amazon and Flickr
These proposed models were evaluated using a small selection of datasets including the citation networks, Reddit, Amazon and Flickr. These datasets are also prevalent in current literature as a method of comparing new GNN architectures against prior architectures. This results in any shortcomings in these datasets propagating through successive papers.
%% These are standard datasets in the field with the need to compare against prior techniques cauting propagation
%% For this reason OGB was also created which has a set of datasets for node classification
To combat this Hu et al. \cite{hu2020open} developed the Open Graph Benchmark to standardise the datasets used for comparison.

% Notice that these approaches are focused on classical text classification techniques

However, we find that the vast majority of datasets focus on text classification tasks and utilise bag of words (BoW) extensively. \textbf{This is not a good representation of all tasks that GNNs may be used for} as BoW is not always applicable to the raw data. Furthermore, simple text classification does not require particularly complex or rich node features and therefore does not test the capabilites of GNNs.

%% All of these datasets focus on simple text classification 
%% These tasks do not utilise complex data structures
%% These tasks also do not require particularly feature rich data

%-> how do they differ in problem seting

%-> how are they applicable? or why are they not?

% Pitfalls of Graph Neural Network Evaluation
% Similarly looking at how effective GNN evaluation is
% Also focus on the dataset
% But they look at the data split of the dataset
% We look at how the dataset embeddings are created maintaing the same split
% Both ideas are important to consider in the influence of datasets
% \textbf{Related Work}
\citet{shchur2018pitfalls} focused on common pitfalls in GNN evaluation. Changing the train, evaluation and test split on a dataset can cause large changes in accuracy and the rank of different GNN architectures, even though other hyperparameters remained constant. This paper supports the idea that there are many factors that might affect the evaluation of the performance of GNNs. However, \citet{shchur2018pitfalls} did not look into the embedding of the data which we find to be important for model performance. 
%To prevent the problems outlined by \citet{shchur2018pitfalls} we maintain the same split and training parameters across datasets.

%\input{sections/background}
\section{Datasets}

%What we did in this paper

% We generate 3 new datasets based on Flickr and Amazon

%In this work we present three new datasets based on publically available Flickr and Amazon data. 

%% To focus on the requirement of alternative techniques we focus on Flickr
%% To maintain the use of text classification we also include two subsets of Amazon
%% In all cases these datasets are node classification texts

We introduce three new datasets: Flickr\_v2, to highlight the importance of embeddings for non-text databases, and two Amazon datasets (AmazonElectronics and AmazonInstruments), to evaluate new text embeddings against the standard bag of words (BoW) approach.
In each dataset we carry out a meta-label approach to generate labels for each node. Each node in the three datasets has a set of tags or categories which are converted into a word vector through GLoVe. These vectors are compared to each meta-label vector and the closest is chosen as the new label.
Review and comparison to other datasets is detailed in \Cref{sec:app:dataset-appendix}.

% Each of these datasets have a selection of embeddings

% These embeddings use classical approaches were applicable and all have neural network approaches

%% In the case of Flickr the standard BoW embedding does not make sense or equates to color histogram
%% Modern image classification networks have shown good accuracy on a range of images
%% Using the resulting feature space before classification should yield a sensible node embedding
%% Importantly we can use pre-trained weights for these image classification networks
%% We use a small selection of available convolutional neural networks with pre-trained weights

\subsection{Flickr\_v2}
This is an image classification task using the same graph structure as the prior Flickr \citep{mcauley2012image} dataset. Each node represents an individual image in the network where the raw data is the image data itself. Each edge represents some connection between images based on comments, likes and groups.
As stated before, \textbf{in the case of images, BoW is not applicable}. Instead using image classification modelss provides a method of encoding raw image data in a compact feature space. Extracting these node embeddings from this feature space yields a sensible embedding to be used in a graph dataset.

We can initialise these image classification models with pre-trained weights to embed the images rather than needing to train the model ourselves. This does also open up the possibility of carry out further training of the image classification model to be better suited to GNN providing adaptable node embeddings. For the Flickr\_v2 we use a selection of convolutional neural networks (CNNs), namely two variations of ResNet \citep{he2016deep} (ResNet18 and ResNet50) and VGG16 \citep{simonyan2015very}, to provide the three embeddings for the Flickr dataset. 
%%% All the images are downloaded in their original dimensions and a pre-transform is applied depending on which network is used.

It is important to highlight the exclusion of BoW from Flickr\_v2 as this is the embedding used in the current Flickr dataset \citep{mcauley2012image}. Given that Flickr\_v2 uses raw image data we cannot sensibly generate BoW embeddings and instead need to use an image based technique. The previous Flickr used human descriptions of the images which is not always possible when all that is available is raw image data.

%% In the case of Amazon BoW is a suitable candidate and thus we include this option
%% As an alternative embedding we use a roBERTa model trained on MNLI
%% This provides three separate embeddings as we have the byte pair encoding of sentences, the roBERTa encoding of sentences, and the final feature space before classification
%% It is important to note that roBERTa is not optimised for simple text classification and further studies with models that are better suited needs to be completed

\subsection{AmazonElectronics and AmazonInstruments}
Both Amazon datasets are text classification tasks using the graph structure induced by the ''similar items", ''co-viewed" and ''co-bought". Each node represents a single item from a the specified category where the raw data is the review text.

In comparison to Flickr\_v2 BoW is a suitable candidate as an embedding as the raw data is text reviews. Keeping with the approach for Flickr\_v2 we utilise a text transformer model, specifically the roBERTa \citep{liu2019roberta} transformer. Compared to the CNNs we have multiple stages in the classification to extract embeddings from: the preprocessing step converting the raw text into byte-pair encodings, the transformer encodings and the final feature vector before classification. These three embeddings are called Byte-Pair, roBERTa-Encoded and roBERTa respectively.
%The way that roBERTa operates is that it takes the raw text and generates a byte-pair encoding of the text as input into the model. We take these byte-pair encodings as one of our text embeddings referred to as \textit{roBERTa Byte-Pair}. After this step the text goes through a block of transformer encoders resulted into a vector encoding the input text, this embedding we refer to as \textit{roBERTa-Encoded}. Finally these encodings are used in a classifier to produce the final classification, this results in a final feature vector before a linear classifier which we use as our final embedding, \textit{roBERTa}.

The roBERTa model also uses pre-trained weights in this case we use a roBERTa model trained on the MNLI dataset. This does mean that this specific model is not optimised for simple text classification, as MNLI is designed for sentiment analysis.
%%% As we are using roBERTa the total number of tokens present in the review text has a limit. This means that a number of items were removed as they had reviews that exceeded the limit of the roBERTa byte-pair encoder.

%\textbf{Comparison to Existing Datasets}

% We aim to use the same graph structure present in prior papers to only change the node embeddings

%% Though we build these datasets from scratch we utilise existing adjacency matrices
%% This means that the bar issues with data no longer being available we use the same graph structure as previous datasets
%% Thus we claim that problems with graph structure would be prevalent in the existing datasets as well and that we can test the change in embedding only

%Why we did what we did

% Important to note that Flickr_v2 does not have bag of word embeddings
% This is an example of a data structure that has no sensible BoW or word vector embedding
% Conversely Amazon does have bag of words as this is a sensible addition

%% The choice of leaving BoW embeddings off from Flickr_v2 is because this approach does not make logical sense here
%% When dealing with non-text-based data we need better embedding techniques
%% We retain BoW in Amazon as a comparison of existing techniques to potentially new text-based techniques

%Build on prior chapters for a full understanding of our approach
%\input{sections/setup}
\section{Evaluation}
\label{sec:eval}

% We use the same GNN architecture across the different datasets

%% All GNN architectures utilise those presented in their respective papers
%% This consists of two layers and a classification layer across all models 
%% We use this same architecture for each dataset, embedding pair

\subsection{Experimental Setup} 
Each GNN is setup following their respective papers, in all cases this entails a 2 layer architecture with a final classification layer. Each of these architectures remains the same across all (dataset, embedding) pairs.
% This is based on the similarity of input size and source data of the new datasets and existing datasets
%% Given the similarity of the graph data and embeddings we utilise the same hyperparameters for the datasets
As the datasets are based on existing datasets where the graph structure and input vector sizes are similar we use the same hyperparameters as the original papers.
To prevent bias from training we use the same optimizer.

% The same optimizer and no learning rate scheduler is used for the GNN training process

%% Keeping the rest of the training process consistent we use the same optimizer
%% We also maintain the same learning rate throughout training

% each dataset is split into train/val/test at 70/10/20 and this is consitent across embeddings

%% For each dataset we create random splits for train, validation and test accompanying 70%, 10%, 20%
%% Each epoch of training uses the train and validation splits whilst retaining test for the final test run

Each dataset is split into train, validation and test splits into 70\%, 10\% and 20\% respectively. The same split is used across all the embeddings for a given dataset to prevent this influencing the performance. Each epoch of training uses the train and validation splits with the test split held out for evaluation. Each GNN is given 300 epochs to train on a (dataset, embedding) pair and we carry out 3 runs to calculate a mean accuracy and confidence interval.

% We take 3 runs for each GNN on each (dataset, embedding) pair and using the same set of 3 random seeds

\begin{table}[!ht]
	%\vspace{-10pt}
	\centering
	\caption{Test accuracy on Flickr\_v2 with different embeddings. }
	\label{tab:flickr-std}
	\begin{tabular}{cccc}
		\toprule
		\multirow{2.5}{*}{Model} & 
		\multicolumn{3}{c}{Embedding Styles} \\
		\cmidrule(lr){2-4} &
		\multicolumn{1}{c}{ResNet18} & 
		\multicolumn{1}{c}{ResNet50} &
		\multicolumn{1}{c}{VGG16} \\
		%\midrule
		%Unconnected Model & 45.2\% $\pm$ 0.1 & \textit{46.9\%} $\pm$ 0.0 & \textbf{47.0 $\pm$ 0.1} \\
		%$\Delta \uparrow$ & +0.0 & \textit{+0.0} & \textbf{+0.0} \\
		\midrule
		GCN & 41.8\% $\pm$ 0.4 & 38.3\% $\pm$ 0.5 & \textbf{{35.5\% $\pm$ 0.3}} \\ % To highlight this as the best GNN model
		%$\Delta \uparrow$ & -3.4 & -8.6 & \underline{\textit{-11.5}} \\                      % given that VGG16 is in bold
		GAT & 38.1\% $\pm$ 0.6 & 37.1\% $\pm$ 1.1 & 27.3\% $\pm$ 1.2 \\
		%$\Delta \uparrow$ & -7.1 & -9.8 & -19.7 \\
		GAT2 & 42.1\% $\pm$ 1.8 & 41.0\% $\pm$ 1.5 & 34.2\% $\pm$ 0.8 \\
		%$\Delta \uparrow$ & -3.1 & -5.9 & -12.8 \\
		GraphSAGE (Random) & 
		\textit{45.4\% $\pm$ 0.1} & 
		\textbf{47.0\% $\pm$ 0.0} & 
		\textit{35.2\% $\pm$ 0.2} \\
		%$\Delta \uparrow$ & \textit{+0.2} & \textbf{+0.1} & -11.8 \\
		GraphSAGE (Neighbour) & 
		\textbf{45.8\% $\pm$ 0.2} & 
		44.5\% $\pm$ 0.1 & 
		34.5\% $\pm$ 0.2 \\
		%$\Delta \uparrow$ &+\textbf{0.6} & -2.4 & -12.5 \\
		\bottomrule
	\end{tabular}
% 	\vspace{-10}
\end{table}

% Looking at Flickr first
% We see that the model that performs best is not consistent across embeddings

%% We see that in the case of ResNet GraphSAGE performs the best though the sampler changes
%% We see that the increase in accuracy is outside of the range of our confidence interval
%% This result goes contrary to the results presented in GraphSAINT when looking at Flickr where GAT2 performed best
%% However, we do see that GAT2 does consistently out-perform GAT
\subsection{Flickr\_v2 and AmazonElectronics Results}
\Cref{tab:flickr-std} demonstrates how the different image node embeddings affect the performance of the five models. We see in the case of ResNet the best performing model is graphSAGE with the only difference being the sampler. This is contrary to the previous results seen on the prior Flickr dataset where Graph Attention Network (GAT) and the improved version GATv2 out-performed GraphSAGE. 

% Though it is important to note that variance between models is far smaller in VGG16

%% Furthermore we see that VGG16 as an embedding performs very badly, but it is important to not that it achieves 47% pm 0.1 as standalone
%% This discrepancy could be due to the size of embedding though more likely due to how ResNet and VGG are designed with ResNet providing richer features
%% It is also important to note that for the majority of models the mean accuracy lies within the confidence interval of the other models
%% This suggests that overall VGG16 does not produce high quality embeddings

When looking at the results for VGG16 we notice that there is less variation in the results for each GNN and a reduction in the accuracy of the models.
%%% this indicates that the feature space VGG16 uses is not well suited for GNNs. 
However, when trained on the underlying images VGG16 out-performed both ResNet18 and ResNet50 achieving 47.0\% accuracy compared to 45.2\% and 46.9\% respectively for the ResNet models. Therefore it is not only the performance of the network used to create the embeddings that is important but rather the feature vectors produced before classification.

% In the cases of ResNet both times GraphSAGE performs best with only the sampling playing a role

%% These results highlight the importance of embedding
%% In this case VGG16 is a good standalone model but a poor source of embeddings
%% Furthermore the ranking of models differs from the BoW approach of GraphSAINT favouring GraphSAGE instead

We also see that the ranking of the models remains relatively consistent across the embeddings, though in the case of VGG16 the models perform relatively the same. Importantly \textbf{this ranking is different from those presented when evaluating on the bag of words (BoW) version of Flickr}.

\begin{table}[!ht]
	\centering
	\caption{Test accuracy on AmazonElectronics with different embeddings.}
	\label{tab:amazon-std}
	\begin{tabular}{ccccc}
		\toprule
		\multirow{2.5}{*}{Model} & 
		\multicolumn{4}{c}{Embedding styles} \\
		\cmidrule(lr){2-5} &
		\multicolumn{1}{c}{Bag of Words} & 
		\multicolumn{1}{c}{Byte Pair} &
		\multicolumn{1}{c}{roBERTa Encoded} & 
		\multicolumn{1}{c}{roBERTa}\\
		%\midrule
		%Unconnected MLP & 71.6\% $\pm$ 0.3 & 21.6\% $\pm$ 0.0 & 55.8\% $\pm$ 0.1 & 51.9\% $\pm$ 0.2 \\
		%$\Delta \uparrow$ & +0.0 & +0.0 & +0.0 & +0.0 \\
		
		\midrule
		GCN & 69.1\% $\pm$ 0.1 & 21.7\% $\pm$ 0.2 & 22.7\% $\pm$ 1.1 & 22.3\% $\pm$ 1.2 \\
		%$\Delta \uparrow$ & -2.5 & +0.1 & -33.1 & -29.6 \\
		
		GAT & \textit{81.1\% $\pm$ 0.2} & 22.2\% $\pm$ 0.5 & 46.1\% $\pm$ 1.5 & 40.3\% $\pm$ 2.9 \\
		%$\Delta \uparrow$ & \textit{+10.5} & +0.6 & -9.7 & -11.6 \\
		
		GAT2 & \textbf{81.8\% $\pm$ 0.3} & 22.2\% $\pm$ 0.6 & 41.8\% $\pm$ 5.1 & 35.7\% $\pm$ 5.6 \\
		%$\Delta \uparrow$ & \textbf{+10.2} & +0.6 & -14.0 & -16.2 \\
		
		GraphSAGE (Random) & 71.3\% $\pm$ 0.1 & \textit{26.3\% $\pm$ 0.3} & \textit{57.0\% $\pm$ 0.5} & \textit{53.7\% $\pm$ 0.5} \\
		%$\Delta \uparrow$ & -0.3 & \textit{+4.7} & \textit{+1.2} & \textit{+1.8} \\
		
		GraphSAGE (Neighbour) & 76.4\% $\pm$ 0.3 & \textbf{40.4\%} $\pm$ 0.4 & \textbf{67.8\% $\pm$ 0.4} & \textbf{66.4\% $\pm$ 0.3} \\
		%$\Delta \uparrow$ & +4.8 & \textbf{+20.8} & \textbf{+12.0} & \textbf{+12.5} \\
		\bottomrule
	\end{tabular}
\end{table}

% Looking at Amazon
% We clearly see that Bag of Words is superior here and roBERTa does not compete

%% These tables represent the 4 different embeddings for the two amazon datasets
%% We clearly see that the best performing model is GAT2 on the Bag of Word embeddings
%% This result follows the results from prior papers including the ranking of the models
%% But when looking at embeddings generated from roBERTa we do see a large drop off
%% However we do see the return of GraphSAGE being the best suited with variation in rank
% \textbf{AmazonElectronics Results}
\Cref{tab:amazon-std} demonstrates how different text node embeddings affects the performance of the models. It is clear that the common standard of Bag of Words (Bow) is far superior to other embedding styles. What is more interesting is that fact that in this case we see that both GAT models out-perform the other models by a significant margin. But when looking at the roBERTa encoding we see that GraphSAGE performs the best in line with what we see in Flickr\_v2.

% But interesting we note that here GAT performs the best -> in line with prior publications

%% It is an important observation that the expected outcome based on prior papers occurs in Bag of Words
%% We see that the newer models perform better on this embedding which is prevalent embedding type
%% Granted this embedding is by far the best, however this can't be applied to Flickr

Unlike in the case of Flickr\_v2 we see that the results are more inline with the results we see from previous BoW datasets. We also see that the ranking of the models follows those that are presented in prior papers suggesting that \textbf{improvement of models has been focusing on optimising for BoW embeddings}. In this case the best embedding option is BoW, however, not always applicable. 

It is important to note that we are not promoting an alternative embedding function other than BoW, since BoW shows the best performance in \Cref{tab:amazon-std}. However, we are showing the phenomenon that \textbf{current GNN architecture design and evaluation is promoting overfitting to BoW embeddings}.
% We see in roBERTa embeddings, though worse, that GraphSAGE is by far the best option
%% When comparing alternative embeddings we find that GraphSAGE frequently out-performs GAT
%% Therefore in a situation where the embeddings are not bag of word or similar discrete embeddings GAT is not necessarily best suited
We see in the majority of node embeddings, which are not BoW, that GraphSAGE performs the best. This suggests that when using a wider range of embeddings our expected accuracy is better if we were to use GraphSAGE over GAT even though GAT is often consider state of the art.
Results for AmazonInstruments are available in \Cref{sec:app:results}.

% It is important to note that in this example we don't claim Bag of Words is ill fitting but that it has a stark influence on performance

%% Flickr_v2 demonstrates there are multiple data where Bag of Words is not suitable
%% In these cases a re-evaluation of what the "best" GNN is needs to be conducted regardless of underlying graph structure
%% Furthermore if a general GNN is to be developed these discrepancies in embedding performance need to be accounted for

% A Repeat that if given raw image data it is not possible to generate Bag of Words

% Note that for a simple text classification task BoW is expected to perform better

%% As an aside in the case of text classification we anticipate BoW performing better
%% The labels are representations of words describing certain categories
%% The words in the label frequently appear in the bag of words
%% A simple system that identifies which part of the BoW corresponds to which category will yield high accuracy
\subsection{Discussion and Limitations}
\textbf{Discussion} In the case of text classification where our labels are representations of words or phrases we anticipate BoW to perform best. This is because BoW provides a discrete collection of word presence, if these can be linked to words in the labels then there is a simple direct connection between a few bits in the node vector and the output classification. In the case of Amazon the BoW vectors contain direct synonyms of the label words or phrases.

% Ideally a mathematical justification for GAT performing best

%% Note this from of discretely picking out features in an input feature vector is ideal for GAT
%% We hypothesise that with more heads than labels each head can focus on identifying which parts of the vector correspond to each category
%% Further studies into more complex text based tasks or labelling is still required to verify this

This invites the question as to why GAT performs far better on BoW wherease GraphSAGE performs better on model embeddings. In the case of BoW we have discrete inclusion or exclusion of a specific word but in the case of model embeddings we have a continuous vector that varies within the model's feature space. Therefore, it is more likely that GAT is good at picking out discrete features than GraphSAGE, which is to be expected given the architecture can utilise multiple heads to focus on individual entries. On the contrary GraphSAGE is better suited to continuous vectors as it takes the whole vector into account at once when computing a classification.

In comparison, the field of NLP is moving away from BoW. This is mainly because more complex tasks such as sentimental analysis or language modelling benefit from richer embeddings 
%in the vector space. 
This in turn questions the focus on text-based graph datasets: \textbf{is text classification on a graph really the task we want to use to assess the quality of our GNNs?} 

% A pondering on how architecture would change if roBERTa or ResNet embeddings were used instead

%% But it is interesting to consider how architecture would change if we used embeddings from existing neural networks

%Compare to the baseline in related work

%Would ideally have ablation studies and this would be required for ICML

% In here we would include hyperparameter search
% In here we would include dataset splits? (but we already see that messes shit up)
% In here we would spend time procrasting on not writing the paper >:(

% This is a small subsection of GNN representation learning tasks but trying to index all options would be unreasonable

%% The results in this section are a small subsection of all the representation learning tasks
%% We only focus on node based embeddings
%% And we focus on two specific datasets rather than exploring a wider range of data
\textbf{Limitations}
The datasets we provide are a small subsection of all possible representation learning tasks that could be carried out on graph networks. Similarly we only provide a handful of embeddings and do not endeavour to find the optimal embedding for each tasks. Thus these results do not represent all possible GNN tasks however we do still see clear trends.
% We do not push the limits of sequence data tasks
%% When discussin text datasets we only look at the simple text classification task
%% Though this is an industry standard pushin this boundary will hopefully yield new GNN architectures
%% These may be more general an better suited to other embeddings
% Further studies into non-standard GNN is not undertaken due to the brevity of this format
%% We also only focus on standard GNNs that use the classic DNN style of stacked layers
%% Multiple approaches to GNN representation learning exist and these likely behave very differently
%% With the success of simplified GNNs such as SGC if these perform consistently across embeddings we may find this approach is better suited than current complex models
Due to the brevity of an extended abstract we only focus on variations on three standard models rather than analysing the effect of embeddings on more niche models. There has been a lot of work into simplified GNNs moving away from the layered approach of the models presented in this paper. These new models may be better suited or more consistent across the different node embeddings.
\section{Conclusion}

%Recap of the problem setting, our method to demonstrate/tackle this, and the results that we see from that

% We see a prevalence of Bag of Words or other classical text classification techniques in GNN datasets

Current approaches to evaluating graph neural networks (GNNs) focuses on text classification using bag of words (BoW) embedding to transform the raw text into a compact node feature.
However, this approach is not general for all types of data and thus the evaluation of GNNs is overfitting to BoW and text classification.
% These approaches are not always suited or possible for the rich data available
%% This approach of embedding is not suited to all types of data
%% Focusing on BoW and text classification means that we are not evaluating against a wide range of tasks
% We instead use modern machine learning approaches to generate node embeddings
%% Introducing image classification in a graph network we utilise CNNs to produce embeddings
%% We also see whether other text classification networks provide better embeddings
% We compare modern machine learning approaches with classical techniques in a limited setting
% We find that the performance of GNNs depends on the node embedding used

Our work demonstrates how the GNN performance is dependent on the node embeddings used in training and evalutation, providing new embedding candidates where BoW is not applicable.
In evaluating different choices of embeddings we introduce three new datasets each with their own set of embeddings. We show that each node embedding favours different GNN architectures rather than simply effecting the accuracy. 
\bibliographystyle{iclr2023_conference}
% ------------------------------------

\bibliography{references}

\newpage
\appendix
\section{Comparison to Other Datasets}
\label{sec:app:dataset-appendix}

We aim to maintain the same graph structure as previous versions of our three datasets describted below. This is to keep the number of varying factors to a minimum and just focus on how the embeddings effect the result. To achieve this we use the same adjacency matrices published with these prior datasets. Although we were unable to source all the raw data these deviations are limited.

Due to the importance of $f_e$ on GNN performance there is a lot to discuss about prior datasets that exist within the space of GNNs in regards to these functions.

% \subsubsection{Torchvision}

% \textbf{UNCERTAIN WHAT TO DISCUSS HERE}

% Though this paper focuses on graphly-connected datasets there is still a discussion to be had about existing datasets outside of graphly-connected ones. It can be argued that these datasets employ some embedding function $f_e$ as the images provided in these datasets need to be scaled and normalised. But these transformations are not embedding functions which is an important distinction.

\subsection{Pytorch Geometric}
\label{sec:app:pyg}

Pytorch Geometric python library provides a standard interface on top of Pytorch to allow for the development of graph based machine learning. The library also provides a sample of datasets from previous papers published in this field.

As is clear from the table the current standard embedding for datasets is bag of words. In the cases where bag of words approaches are not used the approach is grounded in classical text representations such as n-grams and word vectors.

The tasks in these popular datasets are node classification where the node data is frequently text. We therefore say that on the node level these are text classification tasks. The only instance of a non-text classification task is Flickr \cite{zeng2019graphsaint}, though based on the fact that the underlying data is image descriptions this could also be considered a text classification task.

This demonstrates how limited the reach of GNNs currently stand as they are being trained on datasets that behave very similarly where the only difference is the specifics of the available data. We feel that this does not therefore fully test the capabilities of GNNs and puts too much emphasis on bag of words and text classification.

\subsection{Open Graph Benchmark}

The results in this paper focus on \textit{node property prediction} as the data that unconnected models ordinarily work on is easily transferred to nodes in a graph. So when discussing Open Graph Benchmark  \cite{hu2020open} the focus is on the node property prediction subset (OGBN). 

The goal of OGB is to create a standard set of datasets that can be used to compare different GNN architectures so a discussion as to way we did not use their datasets is warranted. The available datasets \textit{ogbn-products, ogbn-proteins, ogbn-arxiv, ogbn-papers100M} and \textit{ogbn-mag} all use variations on the same text representations used in \Cref{sec:app:pyg}. These include Bag of Words (BoW), word2vec and skip-gram. This means the same discussions on these classical text holds here. 

We see that the majority of the tasks focus on text classification, excluding \textit{ogbn-protein}, this again draws into question how well these datasets are testing the range of classification tasks. Further to this, focusing mainly on BoW style embeddings raises the question of whether we are building good BoW extractors or graph information extractors.

\subsection{Flickr}
The prior Flickr dataset used in \textit{Zeng et al.} \cite{zeng2019graphsaint} originated from \textit{McAuley et al.} \cite{mcauley2012image} which aimed to utilize network connections and image descriptions rather than the images themselves. The specific embedding function that the paper used is Bag of Words.
% , where the input dataset $\bm{X}$ is the descriptors of and image and the embedding function $f_e$ is function that takes a descriptor and returns a vector representing which of the top 500 words of the entire dataset $\bm{X}$ were present.

This embedding function is a valid representation of images but it is not easily applicable to other image datasets. Thus GNNs trained on this dataset are confined to images with descriptions that have been transformed using the same top 500 words. Noting that this list of top 500 words is not readily available.

\subsection{Amazon}
\textit{Zeng et al.} \cite{zeng2019graphsaint} also provide an Amazon dataset (AmazonProducts) covering the entirety of Amazon. Without a known source we instead use available Amazon databases online to download and generate our own dataset. The embedding function  used is to tokenise the reviews by 4-grams and take the single value decomposition. This is, as with Flickr, not easily applicable outside of the original dataset.

An alternative Amazon dataset (Amazon) is also available from \textit{Shchur et al.} \cite{shchur2018pitfalls} created originally in \textit{McAuley et al.} \cite{McAuley2016image}. Though the original source of the dataset used a pre-trained Caffe model to embed the product images this dataset did not use these. Instead they created their own embeddings using the bag of words standard with the product reviews as the raw data.
%\input{sections/definitions}
%\input{sections/granet}

% These weren't commented out
%\input{sections/related-datasets}
%\input{sections/datasets}
%\input{sections/appendix-models}
\section{Further Results}
\label{sec:app:results}

\Cref{tab:instruments-std} contains the further results collected for AmazonInstruments

\begin{table}[!ht]
	\centering
	\caption{Test accuracy on AmazonInstruments with different embeddings demonstrating how the different embeddings effects the performance and relative ranking of GNN models. Included is 
	%the difference $\Delta$ of each model to the unconnected MLP and 
	the standard deviation of each result.}
	\label{tab:instruments-std}
	\begin{tabular}{ccccc}
		\toprule
		\multirow{2.5}{*}{Model} & 
		\multicolumn{4}{c}{Embedding Styles} \\
		\cmidrule(lr){2-5} &
		\multicolumn{1}{c}{Bag of Words} & 
		\multicolumn{1}{c}{Byte Pair} &
		\multicolumn{1}{c}{roBERTa Encoded} & 
		\multicolumn{1}{c}{roBERTa} \\
		%\midrule
		%Unconnected MLP & 66.1\% $\pm$ 0.2 & 21.0\% $\pm$ 0.3 & 43.9\% $\pm$ 0.4 & 39.8\% $\pm$ 0.7 \\
		%$\Delta \uparrow$ & +0.0 & +0.0 & +0.0 & +0.0 \\
		
		\midrule
		GCN & 64.0\% $\pm$ 0.5 & 20.8\% $\pm$ 0.3 & 20.4\% $\pm$ 0.8 & 20.4\% $\pm$ 0.8 \\
		%$\Delta \uparrow$ & -2.1 & -0.2 & -23.5 & -19.4 \\
		
		GAT & \textit{79.3\% $\pm$ 0.6} & 21.6\% $\pm$ 0.9 & 47.5\% $\pm$ 1.9 & 46.1\% $\pm$ 4.3 \\
		%$\Delta \uparrow$ & \textit{+13.2} & +0.6 & +3.6 & +6.3 \\
		
		GAT2 & \textbf{79.4\% $\pm$ 0.3} & 21.2\% $\pm$ 0.6 & \textit{49.8\% $\pm$ 5.0} & \textit{47.8\% $\pm$ 2.8} \\
		%$\Delta \uparrow$ & \textbf{+13.3} & +0.2 & \textit{+5.9} & +8.0 \\
		
		GraphSAGE (Random) & 67.5\% $\pm$ 0.3 & \textit{23.9\% $\pm$ 0.6} & 45.1\% $\pm$ 1.2 & 41.9\% $\pm$ 0.6 \\
		%$\Delta \uparrow$ & +1.4 & \textit{+2.9} & +1.2 & +2.1 \\
		
		GraphSAGE (Neighbour) & 72.6\% $\pm$ 0.3 & \textbf{43.4\% $\pm$ 0.5} & \textbf{62.4\% $\pm$ 0.5} & \textbf{59.9\% $\pm$ 0.6} \\
		%$\Delta \uparrow$ & +6.5 & \textbf{+22.8} & \textbf{+18.5} & \textbf{+20.1} \\
		\bottomrule
	\end{tabular}
\end{table}
\section{Hyperparameters}
\label{sec:app:hyperparameters}

\Cref{tab:hyperparameters} details the layers of each model used providing the output hidden features of each layer, the sampler used (the specifics shown in \Cref{tab:samplers}) and the maximum and minimum learning rates. Where there is a difference in learning rates we use a learning rate scheduler that decreases the learning rate when validation accuracy plateaus. Where two models use the same sampler the parameters of those samplers are identical to keep consistency across the tests.

%In the case of GraNet we start with a high learning rate for 20 epochs with the pre-trained model frozen to allow the GNN to train. We then unfreeze the entire model dropping the learning rate sharply and train for another 100 epochs. In the first stage no learning rate scheduler is employed, same as for all GNNs, and in the second stage we apply the learning rate scheduler.

For GraphSAINTSampler all setups use a walk length of 2 with 5 steps sampling 100 nodes per node for normalisation calculation.

\begin{table}[!ht]
	\centering
	\caption{Model architecture, sampler and learning rate}
	\label{tab:hyperparameters}
	\begin{tabular}{ccccc}
		\toprule
		\multirow{2.5}{*}{Model} & 
		\multirow{2.5}{*}{Hidden Features} & 
		\multirow{2.5}{*}{Sampler} & 
		\multicolumn{2}{c}{Learning Rate} \\
		\cmidrule(lr){4-5} & & &
		\multicolumn{1}{c}{Max.} & 
		\multicolumn{1}{c}{Min.} \\
		\midrule
		\multirow{2}{*}{GCN} & 256 & \multirow{2}{*}{Random Node} & \multirow{2}{*}{1e-2} & \multirow{2}{*}{1e-2} \\
		& 256 \\
		
		\cmidrule(lr){2-5} \multirow{2}{*}{GAT} & 256 & \multirow{2}{*}{GraphSAINT RW} & \multirow{2}{*}{1e-2} & \multirow{2}{*}{1e-2} \\
		& 256 \\
		
		\cmidrule(lr){2-5} \multirow{2}{*}{GAT2} & 256 & \multirow{2}{*}{GraphSAINT RW} & \multirow{2}{*}{1e-2} & \multirow{2}{*}{1e-2} \\
		& 256 \\
		
		\cmidrule(lr){2-5} \multirow{2}{*}{GraphSAGE (Random)} & 256 & \multirow{2}{*}{Random Node} & \multirow{2}{*}{1e-3} & \multirow{2}{*}{1e-3} \\
		& 256 \\
		\cmidrule(lr){2-5} \multirow{2}{*}{GraphSAGE (Neighbour)} & 256 & \multirow{2}{*}{Neighbour} & \multirow{2}{*}{1e-3} & \multirow{2}{*}{1e-3} \\
		& 256 \\
		
		\midrule
		%\multirow{3}{*}{MLP} & 256 & \multirow{3}{*}{-} & \multirow{3}{*}{1e-5} & \multirow{3}{*}{5e-7} \\
		%& 256 \\
		%& 128 \\
		
		%\cmidrule(lr){2-5} 
		ResNet18 & \textit{as provided} & - & 1e-4 & 5e-6 \\
		ResNet50 & \textit{as provided} & - & 1e-4 & 5e-6 \\
		VGG16 & \textit{as provided} & - & 1e-4 & 5e-6 \\
		
		%\midrule
		%\multirow{3}{*}{GraNet (MLP + GAT2)} & 256 & \multirow{3}{*}{GraphSAINT RW} & \multirow{3}{*}{1e-2} & \multirow{3}{*}{1e-2} \\
		%& 256 \\
		%& 128 \\
		
		%\cmidrule(lr){2-5} GraNet (ResNet18 + GraphSAGE) & \textit{as provided} & Random Node & 1e-3 & 5e-8 \\
		%GraNet (ResNet50 + GraphSAGE) & \textit{as provided} & Random Node & 1e-3 & 5e-8 \\
		\bottomrule
	\end{tabular}
\end{table}
\begin{table}[!ht]
	\centering
	\caption{Sampler parameters}
	\label{tab:samplers}
	\begin{tabular}{ccccc}
		\toprule
		\multicolumn{1}{c}{Sampler} & \multicolumn{1}{c}{Dataset Split} & \multicolumn{1}{c}{Setup} \\
		\midrule
		\multirow{3}{*}{GraphSAINT RW \citep{zeng2019graphsaint}} & Train & roots: 6000 \\
		& Validation & roots: 1250 \\
		& Test & roots: 2000\\
		\midrule
		\multirow{3}{*}{Random Node} & Train &  \# partitions:512 \\
		& Validation & \# partitions:128 \\
		& Test & \# partitions:256 \\
		\midrule
		\multirow{3}{*}{Neighbour \citep{Hamilton:2017tp}} & Train & \# neighbours:[25, 10], batch size:512  \\
		& Validation & \# neighbours:[25, 10], batch size:128 \\
		& Test & \# neighbours:[25, 10], batch size:256 \\
		\bottomrule
	\end{tabular}
\end{table}

\end{document}